\pdfoutput=1

\documentclass[11pt]{article}

\usepackage[]{emnlp2021}

\usepackage{times}
\usepackage{latexsym}

\usepackage[T1]{fontenc}

\usepackage[utf8]{inputenc}

\usepackage{microtype}

\usepackage{todonotes}
\usepackage{multirow}
\usepackage{graphicx}
\usepackage{enumitem}
\usepackage{anyfontsize}
\usepackage{url}
\usepackage{subcaption}
\usepackage{xr}
\usepackage{tabu}
\usepackage{bbold}
\usepackage{booktabs}
\usepackage{xcolor,pifont}
\usepackage{MnSymbol,wasysym}
\usepackage{fontawesome}
\usepackage{xspace}
\usepackage{caption}
\usepackage{array}
\newcolumntype{H}{>{\setbox0=\hbox\bgroup}c<{\egroup}@{}}

\newcommand{\Skip}[1]{}

\newcommand{\sysname}{\textsc{HyperExpan}\xspace}
\newcommand{\modelname}{\textsc{HyperExpan}\xspace}

\newcommand{\ie}{\textit{i}.\textit{e}.\ }
\newcommand{\eg}{\textit{e}.\textit{g}.\ }
\newcommand{\mobius}{M{\"o}bius\xspace}
\newcommand{\poincare}{Poincar\'e\xspace}

\newcommand{\secref}[1]{\S\ \ref{#1}}
\newcommand{\appref}[1]{Appendix \ref{#1}}
\newcommand{\figref}[1]{Figure \ref{#1}}
\newcommand{\tbref}[1]{Table \ref{#1}}

\newcommand{\dotieconcat}[2]{
  \text{\raisebox{.8ex}{$\smallfrown$}}%
}
\newcommand{\mypar}[1]{\noindent\textbf{#1}}
\newcommand{\wordnetverbARBORIST}
{& 608.7 & 0.280 & 10.8 & 24.0 & 27.7 & 6.7 & 4.8 & 3.2}
\newcommand{\wordnetverbTMN}
{& 465.0 & 0.479 & 14.9 & 31.6 & 37.9 & 13.2 & 6.4 & 4.0}

\newcommand{\wordnetverbTaxoExpanGCN}
{& 502.8 & 0.439 & 12.4 & 28.2 & 35.2 & 12.4 & 5.6 & 3.5}

\newcommand{\wordnetverbShallow}
{& 5495.4 & 0.023 & 1.6 & 1.6 & 1.7 & 1.6 & 0.3 & 0.2}
\newcommand{\wordnetverbRoberta}
{& 2132.8 & 0.172 & 4.3 & 10.1 & 12.6 & 4.3 & 2.0 & 1.3}
\newcommand{\wordnetverbRobertaFT}
{& 1535.7 & 0.155 & 2.4 & 6.4 & 9.9 & 2.4 & 1.3 & 1.0}

\newcommand{\wordnetverbEuclideanMLP}
{& 747.2 & 0.399 & 10.8 & 24.0 & 31.7 & 10.8 & 4.8 & 3.2}

\newcommand{\wordnetverbHyperbolicMLP}
{& 617.4 & 0.419 & 10.5 & 25.6 & 33.7 & 10.5 & 5.1 & 3.4 }

\newcommand{\wordnetverbGCN}
{& 456.9 & 0.445 & 10.9 & 27.2 & 34.5 & 10.9 & 5.4 & 3.5}
\newcommand{\wordnetverbGAT}
{& 471.7 & 0.449 & 11.6 & 28.7 & 35.6 & 11.6 & 5.7 & 3.6}

\newcommand{\wordnetverbOurs}
{& 400.8 & 0.517 & 15.0 & 32.8 & 42.7 & 15.0 & 6.6 & 4.3 }

\newcommand{\wordnetverbFullEuclidean}
{& 450.0 & 0.483 & 13.7 & 31.8 & 39.4 & 13.7 & 6.4 & 3.9}

\newcommand{\wordnetverbOursBold}
{& \textbf{400.8} & \textbf{0.517} & \textbf{15.0} & \textbf{32.8} & \textbf{42.7} & \textbf{15.0} & \textbf{6.6} & \textbf{4.3} }

\newcommand{\wordnetverbPrevAblationBaseline}
{ & 435.6 & 0.506 & 15.0 & 32.8 & 42.2 & 15.0 & 6.6 & 4.2}

\newcommand{\wordnetverbCurvefixBest}
{& 416.7 & 0.490 & 14.4 & 31.7 & 40.8 & 14.4 & 6.3 & 4.1} 

\newcommand{\wordnetverbfasttext}
{& 479.1 & 0.494 & 15.2 & 32.5 & 41.0 & 15.2 & 6.5 & 4.1}

\newcommand{\wordnetverbOursAbsPos}
{& 444.8 & 0.497 & 13.0 & 31.0 & 40.8 & 13.0 & 6.2 & 4.1}

\newcommand{\wordnetverbOursRelPos}
{& 468.7 & 0.503 & 14.3 & 33.4 & 41.2 & 14.3 & 6.7 & 4.1}

\newcommand{\wordnetverbOursNoPos}
{& 466.6 & 0.482 & 12.5 & 30.2 & 38.8 & 12.5 & 6.0 & 3.9}

\newcommand{\wordnetverbOursOneHopAncestors}
{& 446.7 & 0.505 & 15.5 & 33.6 & 42.5 & 15.5 & 6.7 & 4.3}

\newcommand{\wordnetverbOursOneHop}
{& 435.6 & 0.506 & 15.0 & 32.8 & 42.2 & 15.0 & 6.6 & 4.2}

\newcommand{\wordnetverbOursTwoHops}
{& 422.2 & 0.502 & 14.5 & 32.1 & 41.7 & 14.5 & 6.4 & 4.2}
\newcommand{\wordnetnounOurs}
{& 573.6 & \textbf{0.607} & \textbf{23.9} & \textbf{42.1} & \textbf{52.5} & \textbf{24.4} & \textbf{8.6} & \textbf{5.4}}

\newcommand{\wordnetnounGCN}
{& 684.1 & 0.563 & 20.9 & 39.8 & 47.3 & 21.3 & 8.1 & 4.8}

\newcommand{\wordnetnounGAT}
{& 640.7 & 0.585 & 22.3 & 40.9 & 49.7 & 22.7 & 8.3 & 5.1}

\newcommand{\wordnetnounEuclideanMLP}
{&}

\newcommand{\wordnetnounHyperbolicMLP}
{& 869.6 & 0.502 & 18.1 & 33.6 & 41.7 & 18.5 & 6.9 & 4.3}

\newcommand{\wordnetnounARBORIST}
{& 1095.1 & 0.435 & 16.5 & 28.4 & 34.1 & 16.8 & 5.8 & 3.5}

\newcommand{\wordnetnounTaxoExpanGCN}
{& 649.6 & 0.562 & 19.7 & 38.2 & 47.4 & 20.1 & 7.8 & 4.8}

\newcommand{\wordnetnounTMN}
{& \textbf{501.0} & 0.595 & 20.7 & 40.5 & 50.1 & 21.1 & 8.3 & 5.1}

\newcommand{\wordnetnounRoberta}
{& 25235.4 & 0.158 & 13.7 & 15.7 & 15.7 & 14.0 & 3.2 & 1.6}

\newcommand{\wordnetnounRobertaFT}
{& 27748.2 & 0.148 & 5.9 & 13.7 & 13.7 & 6.0 & 2.8 & 1.4}
\newcommand{\magpsyOurs}
{& \textbf{38.4} & \textbf{0.827} & \textbf{28.8} & \textbf{63.0} & \textbf{75.3} & \textbf{37.2} & \textbf{16.3} & \textbf{9.7}}

\newcommand{\magpsyGCN}
{& 51.4 & 0.742 & 23.8 & 52.5 & 64.3 & 30.8 & 13.6 & 7.4}

\newcommand{\magpsyGAT}
{& 48.6 & 0.751 & 23.6 & 52.4 & 65.8 & 30.5 & 13.5 & 8.5}

\newcommand{\magpsySLP}
{& 98.1 & 0.701 & 19.1 & 46.6 & 59.9 & 24.7 & 12.1 & 7.7}

\newcommand{\magpsyHyperbolicMLP}
{& 74.1 & 0.739 & 21.8 & 51.4 & 64.9 & 28.2 & 13.3 & 8.4}

\newcommand{\magpsyARBORIST}
{& 119.9 & 0.722 & 21.0 & 48.4 & 62.9 & 25.8 & 12.5 & 7.7}

\newcommand{\magpsyTaxoExpan}
{& 68.5 & 0.775 & 26.1 & 56.9 & 69.5 & 33.8 & 14.7 & 9.0 }

\newcommand{\magpsyTMN}
{& 73.0 & 0.781 & 25.8 & 58.7 & 70.5 & 33.4 & 15.2 & 9.1 }
\newcommand{\magcsOurs}
{& \textbf{74.4} & \textbf{0.689} & \textbf{16.1} & \textbf{44.6} & \textbf{58.0} & \textbf{26.1} & \textbf{14.5} & \textbf{9.4}}

\newcommand{\magcsGCN}
{& 90.3 & 0.653 & 14.5 & 39.6 & 53.3 & 23.6 & 12.9 & 8.7}

\newcommand{\magcsGAT}
{& 92.2 & 0.676 & 15.9 & 41.9 & 56.0 & 25.9 & 13.6 & 9.1}

\newcommand{\magcsSLP}
{& 136.2 & 0.639 & 16.1 & 41.2 & 52.4 & 26.1 & 13.4 & 8.5}

\newcommand{\magcsHyperbolicMLP}
{& 101.4 & 0.650 & 13.7 & 38.0 & 53.4 & 22.3 & 12.4 & 8.7}

\newcommand{\magcsARBORIST}
{& 284.7 & 0.602 & 15.1 & 38.9 & 49.4 & 24.6 & 12.6 & 8.0}

\newcommand{\magcsTaxoExpan}
{& 189.8 & 0.661 & 15.9 & 42.9 & 55.4 & 25.8 & 13.9 & 9.0}

\newcommand{\magcsTMN}
{& 160.5 & 0.667 & 16.0 & 43.1 & 56.3 & 26.0 & 14.0 & 9.1 }
\newcommand{\SideNote}[2]{\todo[color=#1,size=\small]{#2}} 

\newcommand{\sderek}[1]{\SideNote{green!40}{#1 --derek}}

\newcommand{\muhao}[1]{{\color{blue}[{\sc MC:} #1]}}

\title{HyperExpan: Taxonomy Expansion with \\ Hyperbolic Representation Learning}

\author{
    Mingyu Derek Ma$^1$\quad
    Muhao Chen$^2$\thanks{~~Equal contributions.}\quad
    Te-Lin Wu$^1$\footnotemark[1]\quad
    Nanyun Peng$^{1}$
    \\
    $^1$ 
    Computer Science Department, 
    University of California, Los Angeles
    \\
    $^2$ 
    Department of Computer Science and Information Sciences Institute, \\
    University of Southern California\\
    {\tt \{ma,telinwu,violetpeng\}@cs.ucla.edu}~~~
    {\tt muhaoche@usc.edu}
}

\begin{document}
\maketitle
\begin{abstract}
    Taxonomies are valuable resources for many applications, but the limited coverage due to the expensive manual curation process hinders their general applicability.
Prior works attempt to automatically expand existing taxonomies to improve their coverage by learning concept embeddings in Euclidean space, while taxonomies, inherently hierarchical, more naturally align with the geometric properties of a hyperbolic space.
In this paper, we present HyperExpan, a taxonomy expansion algorithm that seeks to preserve the structure of a taxonomy in a more expressive hyperbolic embedding space and learn to represent concepts and their relations with a Hyperbolic Graph Neural Network (HGNN).
Specifically, HyperExpan leverages position embeddings to exploit the structure of the existing taxonomies, and characterizes the concept profile information to support the inference on unseen concepts during training.
Experiments show that our proposed HyperExpan outperforms baseline models with representation learning in a Euclidean feature space and achieves state-of-the-art performance on the taxonomy expansion benchmarks. 
\end{abstract}

\section{Introduction}
\label{sec:intro}

Taxonomy, a systematic categorization scheme, is an effective way to organize and classify knowledge~\cite{harlin1998taxonomy, stewart2008building}.
Taxonomies have been used to support many downstream applications such as content management in e-commerce~\cite{wang2021enquire,zhang2014taxonomy}, web search~\cite{yin2010building,Liu2019AUC}, digital libraries~\cite{yu2020steam}, and NLP tasks~\cite{yang2020co,hua2016understand,yang2017efficiently}.
The curation of taxonomies mostly relies on human experts, 
which can be time-consuming and expensive, and hence suffer from limited coverage of the knowledge \cite{jurgens2016semeval}.
To alleviate this issue and handle constantly emerging new concepts, automating the taxonomy construction has attracted attentions from the research community~\cite{wang2017short}.
One type of such automated taxonomy curation is \textit{taxonomy expansion}, which enriches an existing taxonomy to incorporate new and broader concepts.
Specifically, the expansion of a taxonomy is performed as attaching new concept nodes to proper positions of a \textit{seed} taxonomy graph, which is usually represented as a hierarchical tree~\cite{vedula2018enriching}.

\begin{figure}[t!]
    \centering
    \includegraphics[width=\columnwidth]{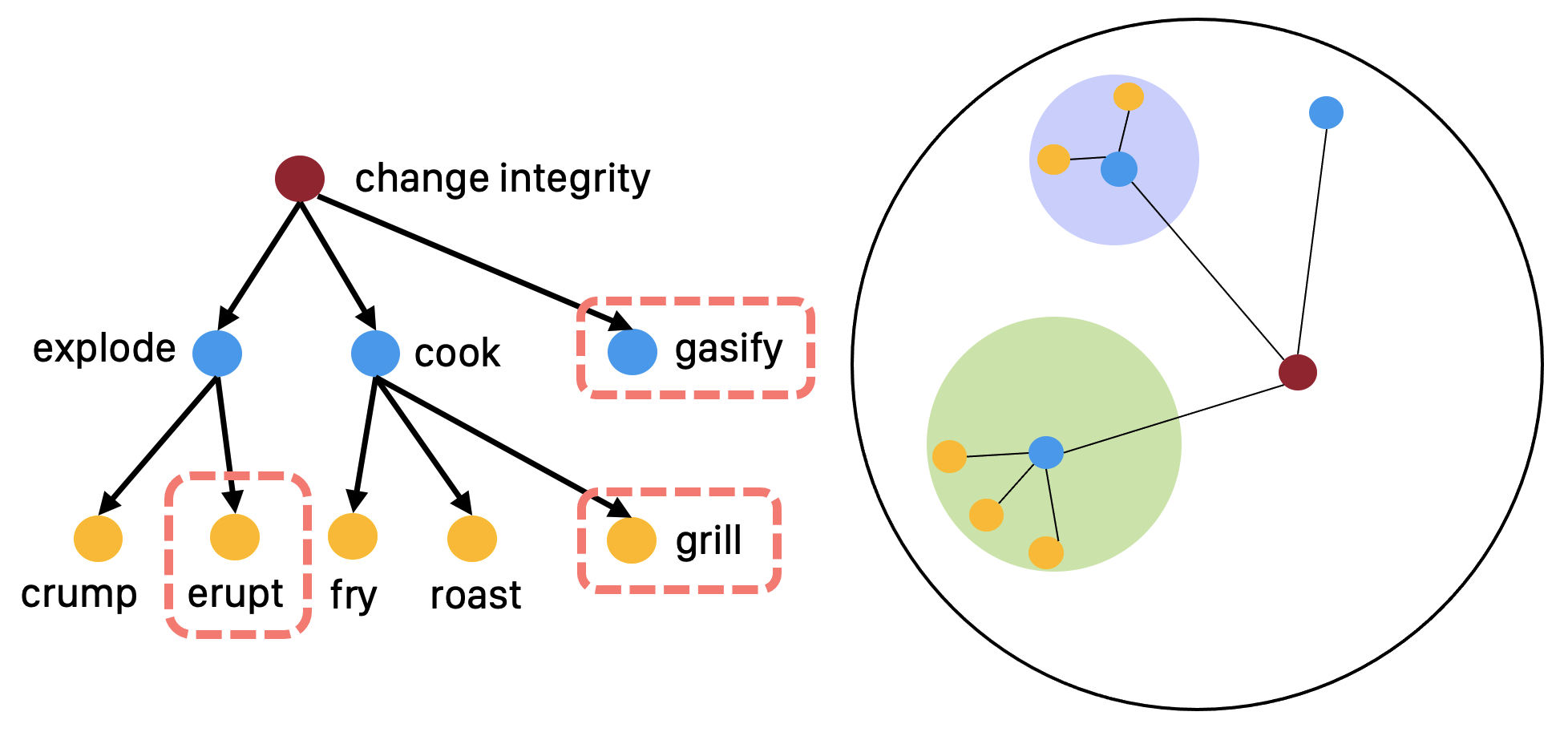}
    \caption{We show the taxonomy expansion task where red boxed concepts are newly attached concepts (left), and illustrate the representation of this taxonomy in a 2D \poincare ball (right). Note that all the black edges have identical hyperbolic lengths.
    }
    \label{fig:teaser}
\end{figure}

To systematically enrich a taxonomy graph, concept embeddings are firstly learned by structurally characterizing the concepts in the existing taxonomies, which are then used to match the embeddings of query concepts for the expansion.
Prior works learn the concept embeddings with local structural features, such as edge semantic representation~\cite{manzoor2020expanding} and graph neural networks (GNN)~\cite{Shen2020TaxoExpanST}.
However, as a concept can lead to multiple subconcepts, the sizes of taxonomies expand exponentially with respect to their levels.
The Euclidean embedding space, where existing works commonly build upon, fails to account for this property.
In contrast, a \textit{hyperbolic space}~\cite{nickel2017poincare, sarkar2011low}, where the circumference of a negative-curved space grows exponentially with regard to the radius as illustrated in~\figref{fig:teaser}, can better capture such special characteristics of taxonomies.

In this paper, we present \modelname, a taxonomy expansion framework based on hyperbolic representation learning, that:
(1) better preserves the taxonomical structure in a more expressive hyperbolic space,
(2) effectively characterizes concepts by exploiting sparse neighborhood information beyond standard parent-child relations~\cite{alyetal2019every,le2019inferring},
and (3) improves inference precision and generalizability by leveraging pretrained distributional features.

Specifically, \modelname incorporates two types of features to exploit the structural presentation of a taxonomy:
a relative positional embedding of a node depending on its relation to the anchor node,
and an absolute positional embedding defined by its depth within a taxonomy.
\modelname  first constructs an \textit{ego subgraph} around the potential attaching candidate concepts, \ie the anchor concepts, and then leverages a hyperbolic graph neural network (HGNN) to obtain the \textit{anchor concept embeddings}.
A parent-child matching score for the attachment is subsequently produced by comparing both the anchor and query concept embeddings in the same hyperbolic space.



We evaluate~\modelname on WordNet and Microsoft Academic Graph datasets. Experiments show that the learned hyperbolic concept embeddings achieve better expansion performance than the Euclidean counterpart, outperforming the state-of-the-art models. 
We also perform ablation studies to demonstrate the effectiveness of each component and the design choice of \modelname.
Our contributions are summarized as follows:
(1) We present an effective and generalizable taxonomy expansion framework via hyperbolic representation learning.
(2) We introduce methods to incorporate pretrained distributional features and taxonomy-specific information in the hyperbolic GNN design.
(3) We show that our framework achieves state-of-the-art performance on expanding four large real-world taxonomies.

\section{Preliminaries}

We introduce preliminaries about hyperbolic geometry and then define the task.

\begin{table*}[t!]
    \centering
    \setlength{\tabcolsep}{3pt}
    {\small
    \begin{tabular}{l|c|c}
    \toprule
         & \poincare Ball & Lorentz Model \\ \midrule
        Distance & 
$
d(x, y)=\cosh ^{-1}\left(1+2 \frac{\|x-y\|^{2}}{\left(1-\|x\|^{2}\right)\left(1-\|y\|^{2}\right)}\right)
$
& 
$d(x, y)=\operatorname{arcosh}\left(-<x, y>_{\mathrm{L}}\right)$
\\ \midrule
        Exponential Map &
$\exp_{x}(v)=x \oplus\left(\tanh \left(\frac{\lambda_{x}\|v\|}{2}\right) \frac{v}{\|v\|}\right)$ 
&
$\exp _{x}^{K}(v)=\cosh \left(\frac{\|v\|_{\mathcal{L}}}{\sqrt{K}}\right) x+\sqrt{K} \sinh \left(\frac{\|v\|_{\mathcal{L}}}{\sqrt{K}}\right) \frac{v}{\|v\|_{\mathcal{L}}}$
\\ \midrule
        Logarithmic Map &
$\log _{x}(y)=\frac{2}{\lambda_{x}} \operatorname{artanh}(\|-x \oplus y\|) \frac{-x \oplus y}{\|-x \oplus y\|}$   
&
$\log _{x}^{K}(y)=d_{\mathcal{L}}^{K}(x, y) \frac{y+\frac{1}{K}\langle x, y\rangle_{\mathcal{L}} x}{\left\|y+\frac{1}{K}\langle x, y\rangle_{\mathcal{L}} x\right\|_{\mathcal{L}}}$
\\ \midrule
    Addition & 
$x \oplus y=\frac{\left(1+2\langle x, y\rangle+\|y\|^{2}\right) x+\left(1-\|x\|^{2}\right) y}{1+2\langle x, y\rangle+\|x\|^{2}\|y\|^{2}}$
& 
$x^{H} \oplus^{K} y:=\exp _{x^{\mathrm{H}}}^{K}\left(P_{\mathbf{o} \rightarrow x^{H}}^{K}(y)\right)$
\\ \midrule
    Matrix Multiplication &
$M \otimes x=\tanh \left(\frac{\|M x\|}{\|x\|} \tanh ^{-1}(\|x\|)\right) \frac{M x}{\|M x\|}$
&
$M \otimes^{K} x^{H}:=\exp _{\mathbf{o}}^{K}\left(M \log _{\mathbf{o}}^{K}\left(x^{H}\right)\right)$ \\
\bottomrule
    \end{tabular}
    }
    \caption{
    Distance metrics and arithmetic operations in  \poincare and Lorentz models.}
    \label{tab:hyperbolic-operations}
\end{table*}

\subsection{Hyperbolic Geometry}
\label{sec:hyperbolic-operations}

Hyperbolic space is a non-linear space with constant negative curvature as opposed to Euclidean space which has zero curvature. The curvature of a space measures how a geometric object deviates from a flat plane.\footnote{Here we assume a unit hyperbolic space (curvature $=-1$) in this section.}
Specifically in this work, we mainly employ the following two models of hyperbolic geometry~\cite{beltrami1868teoria,cannon1997hyperbolic}: the \poincare ball model and the Lorentz model, with some intermediate projective operations defined by the Klein model (see~\secref{subsec:initial-concept-features}).


There are several essential vector operations required for learning embeddings in a hyperbolic space, including:
(1) computing the distance between two points,
(2) projecting from a hyperbolic space to a Euclidean space, and vice versa,
(3) adding and multiplying matrices,
(4) concatenating two vectors, and
(5) transformation among hyperbolic models.
These necessary algebraic operations are summarized in~\tbref{tab:hyperbolic-operations}.

For each point $x \in \mathcal{H}^{n}$ in the hyperbolic space, we denote the associated tangent space centered around $x$ as $T_{x} \mathcal{H}^{n}$, which is always a subset of the Euclidean space.
We make use of the \emph{exponential map} $\exp _{x}: T_{x} \mathcal{H}^{n} \rightarrow \mathcal{H}^{n}$ and \emph{logarithmic map} $\log _{x}: \mathcal{H}^{n} \rightarrow T_{x} \mathcal{H}^{n}$ to project points in the hyperbolic space to the local tangent space for precise approximation, and vice-versa.
Setting the origin (north pole) of the hyperbolic space as the center, we can obtain a common tangent space across different manifolds as long as they are of the same dimension and modeled by the same hyperbolic model using $\log _{0}$ and $\exp _{0}$ projection.
And hence, we can use $\log$ and $\exp$ to perform the projection within a neural network that has a mixture of hyperbolic and Euclidean layers.




The addition and matrix multiplication operations in \poincare model are based on \mobius transformation~\cite{ungar2001hyperbolic,Ganea2018HyperbolicNN,gulcehre2018hyperbolic}, which are defined in~\tbref{tab:hyperbolic-operations}.
In the Lorentz model, we utilize the tangent space to perform matrix multiplication and parallel transport to perform the addition~\cite{Chami2019HyperbolicGC}.

For concatenating two hyperbolic vectors, we perform a generalized version of the concatenation operation~\cite{Ganea2018HyperbolicNN,lopez-strube-2020-fully} to prevent the resulting vector from being out of the manifold, as shown below:
\begin{equation*}
\operatorname{concat}(\mathbf{x}_{1}, \mathbf{x}_{2})=M_{1} \otimes \mathbf{x}_{1} \oplus M_{2} \otimes \mathbf{x}_{2} \oplus b
\end{equation*}
where $M_{1}$, $M_{2}$ and $b$ are parameters.

The \poincare ball model $\mathcal{B}$, the Klein model $\mathcal{K}$ and the hyperboloid/Lorentz model $\mathcal{L}$ are used in our work, and we perform different computation on different models. These models are isometric isomorphic. Given a node $\boldsymbol{x}=\left[x_{0}, x_{1}, \cdots, x_{n}\right] \in \mathcal{L}$, the bijections between node on Lorentz model and its corresponding mapped node on \poincare ball $\boldsymbol{b}=\left[b_{0}, b_{1}, \cdots, b_{n-1}\right] \in \mathcal{B}$ are \cite{cannon1997hyperbolic,iversen1992hyperbolic}:

\begin{equation*}
\begin{aligned}
p_{\mathcal{L} \rightarrow \mathcal{B}}(\boldsymbol{x})&=\frac{\left[x_{1}, \cdots, x_{n}\right]}{x_{0}+1}
\\
p_{\mathcal{B} \rightarrow \mathcal{L}}(\boldsymbol{b})&=\frac{\left[1+\|\boldsymbol{b}\|^{2}, 2 \boldsymbol{b}\right]}{1-\|\boldsymbol{b}\|^{2}}
\end{aligned}
\end{equation*}

The bijections between $\boldsymbol{x}$ and its mapped node on the Klein model $\boldsymbol{k}=\left[k_{0}, k_{1}, \cdots, k_{n-1}\right] \in \mathcal{K}$ are:

\begin{equation*}
\begin{aligned}
p_{\mathcal{L} \rightarrow \mathcal{K}}(\boldsymbol{x})&=\frac{\left[x_{1}, \cdots, x_{n}\right]}{x_{0}}
\\
p_{\mathcal{K} \rightarrow \mathcal{L}}(\boldsymbol{k})&=\frac{1}{\sqrt{1-\|\boldsymbol{k}\|^{2}}}[1, \boldsymbol{k}]
\end{aligned}
\end{equation*}

\subsection{Taxonomy Expansion}
\label{sec:problem-def}

In this work, a taxonomy is mathematically defined as a directed acyclic concept graph $\mathcal{T}=(\mathcal{N},\mathcal{E})$, where each node $n \in \mathcal{N}$ represents a concept, and each directed edge $n_p \rightarrow n_c \in \mathcal{E}$ denotes a parent-child relation in which $n_p$ and $n_c$ is a pair of hierarchically related concepts (\eg \texttt{change integrity} $\rightarrow$ \texttt{explode}). 
Given an existing taxonomy $\mathcal{T}^0=(\mathcal{N}^0, \mathcal{E}^0)$, the goal of the taxonomy expansion is to attach a set of new concepts $\mathcal{C}$ to $\mathcal{T}^0$, \textit{expanding} it to $(\mathcal{N}^0 \cup \mathcal{C}, \mathcal{E}^0 \cup \mathcal{R})$ where $\mathcal{R}$ are new edges whose children must be $c \in \mathcal{C}$.

An illustration of the taxonomy expansion is as shown in~\figref{fig:teaser}, 
where the query nodes (new concepts) are \textit{attached} to the proper positions depending on the surrounding \textit{anchor} nodes (existing concepts).
Following the settings of prior works~\cite{Shen2020TaxoExpanST, zhang2021tmn}, we consider attaching different query concepts independently from each other to simplify the problem.
Each concept in $\mathcal{N}^0 \cup \mathcal{C}$ has its profile information, \ie concept definitions, concept names, and related articles etc. (See~\secref{subsec:datasets} for more details.)

\section{\modelname}

\label{sec:model-design}
\begin{figure*}[th]
    \centering
    \includegraphics[width=1\textwidth]{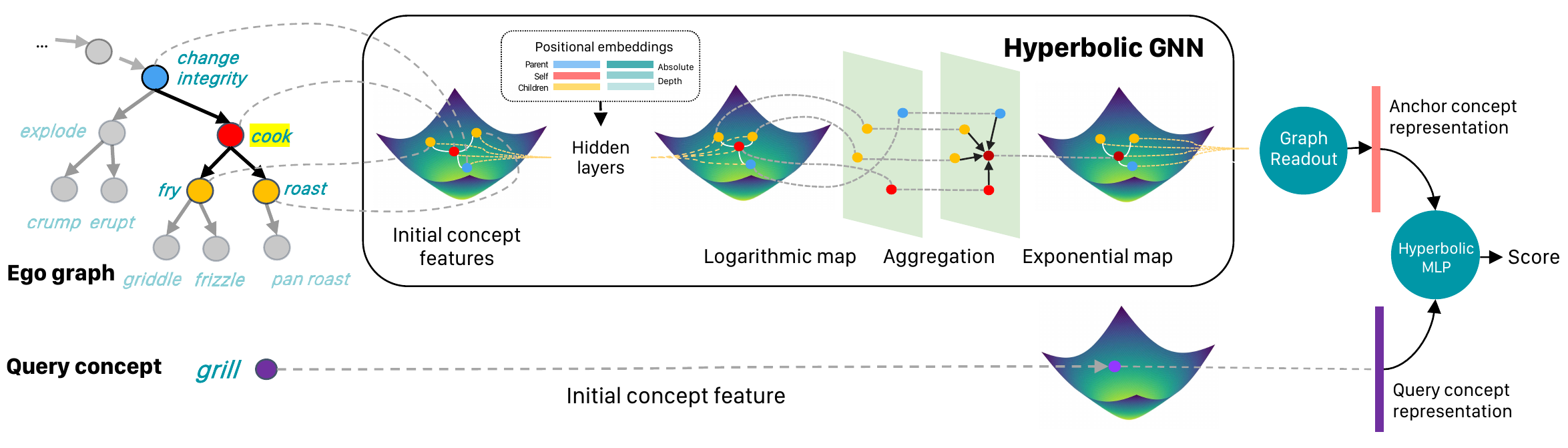}
    \caption{
    \modelname's model design. Red node is the anchor concept and the highlighted sub-tree is the ego graph of the anchor node. The intermediate flat surface is the tangent space based on the anchor node.}
    \label{fig:model}
\end{figure*}

We propose~\modelname, a taxonomy expansion framework based on hyperbolic geometry and GNNs.
As shown in~\figref{fig:model},~\modelname consists of the following main steps:
1) initial concept feature generations utilizing the profile information (\secref{subsec:initial-concept-features}).
2) encoding query and anchor concept features with hyperbolic (graph) neural networks (\secref{subsec:concept-rep}).
3) computing the query-anchor embedding matching scores for attaching query concepts to proper anchor positions (\secref{subsec:match-function}).
We will describe each step in details and how to train the matching model (\secref{subsec:training-learning}) in the following sections.

\subsection{Initial Concept Features}
\label{subsec:initial-concept-features}
We mainly leverage two types of profile information to obtain the initial concept (either in query or existing taxonomy) features: the name and the definition sentences of a concept.
We firstly embed the two profile information by applying an average pooling over the word embeddings of each profile word, and then take the mean of the two embedded profile information to produce the fixed-dimension initial concept embedding.
Our framework does not require the initial word embeddings to be defined in a specific geometry, and thus it can be either Euclidean, such as fastText~\cite{bojanowski2016enriching}, or hyperbolic, such as \poincare~GloVe~\cite{tifrea2018poincare}, which embeds words in a Cartesian product of hyperbolic spaces.
Note that since~\poincare~GloVe is defined in hyperbolic space, the aforementioned mean operation can no longer be the usual Euclidean average since it may produce results that are out of the manifold.
Instead, we use Einstein midpoint method \cite{gulcehre2018hyperbolic} to perform the average pooling.
Denote the token embeddings as $e_i$ and $N$ as number of tokens in a sentence, the midpoint can be computed as:
\begin{equation*}
\mu=\frac{\sum_{i=1}^{N} \gamma_{i} e_i}{\sum_{i=1}^{N} \gamma_{i}}
\end{equation*}
where $\gamma_{i}=\frac{1}{\left\|x_i\right\|^{2}}$ denotes the Lorentz factors.
Einstein midpoint has the most concise form with the Klein coordinates \cite{gulcehre2018hyperbolic}, therefore we project \poincare embeddings to the Klein model $\mathcal{K}$ to calculate the midpoint, and then project the results back to the \poincare model. We project the initial concept embeddings to the hyperbolic space $\mathcal{H}$ initialized by the following network design and used as the network input. 


\subsection{Anchor Concept Representation}
\label{subsec:concept-rep}
We learn a parameterized model to encode anchor nodes $a_{i}$, taking the initial concept features $x_{a_i}$ as inputs, and output the hyperbolic embedding vectors $\mathbf{o}_{a_i}$.
We use HGNN to model the concepts in a hyperbolic space and exploit the structured representation of a taxonomy. We leave the basics of Euclidean Graph Convolutional Networks in~\appref{appendix:gcn}.



HGNN performs the neighbor aggregation operation in a hyperbolic space $\mathcal{H}$, which can be a Lorentz model $\mathcal{L}$ or a \poincare model $\mathcal{B}$, following corresponding numerical operations defined in~\secref{sec:hyperbolic-operations}.
Note that the standard neighbor aggregation operation in (Euclidean) GNN may lead to manifold distortion when embedding graphs with scale-free or hierarchical structure~\cite{deza2009geometry,bachmann2020constant}.

The first layer of an HGNN maps initial node features (can be on a Euclidean or any hyperbolic spaces) to $\mathcal{H}$, followed by a series of cascaded HGNN layers.
At each layer, the HGNN performs four operations in the following order:
1) transforming node features to messages in a predefined hyperbolic space,
2) transforming messages to the tangent space for each node,
3) performing neighborhood aggregation on the tangent space,
and 4) projecting updated tangential node embeddings to hyperbolic space $\mathcal{H}$.
In this work, our HGNN design is based on the hyperbolic graph convolutional network~\cite{Chami2019HyperbolicGC}.

\vspace{.3em}

\mypar{Ego Graph.}
To encode anchor concepts with an HGNN, an \textit{ego graph} centered around anchor concept $a_i$ is firstly constructed, where all nodes on such a graph is bounded by a certain edge distance.

\vspace{.3em}

\mypar{Positional Features.}
To further exploit the structural presentation of a taxonomy, we incorporate the relative and absolute positional embeddings as inputs to an HGNN layer.
With respect to a given center node, the neighbors of such node can be of one of the following three relative positions: parent, child, and self.
Denote $pr_{c}(i)$ as the relative position of node $i$ if the center node is $c$, we have the relative positional embedding as: $\mathbf{x}_{pr_{c}(i)}$.

Motivated by~\citet{you2019position,wang-etal-2019-self}, we equip the HGNN model with the position-awareness by incorporating an absolute position embedding.
We define an \textit{absolute position}, $pa(i)$, of a node $i$ as its \textit{depth} (\ie level w.r.t the root) within the entire taxonomy.
Since the expansion task does not break the structure of the existing taxonomy, such position encoding is fixed for a given node.
The depth-dependent position embedding is defined as $\mathbf{x}_{pa(i)}$.
And hence, the overall inputs to each HGNN layer is a concatenation of the original node embeddings and the two taxonomy-specific features \footnote{Superscript $^\mathcal{E}$ and $^\mathcal{H}$ indicate the node feature is in Euclidean space and hyperbolic space respectively.}:
\begin{equation*}
\mathbf{x}_{i}^{\ell, \mathcal{H}} \leftarrow \mathbf{x}_{i}^{\ell, \mathcal{H}}\ ||\ \mathbf{x}_{pr_{c}(i)}^{\mathcal{H}}\ ||\ \mathbf{x}_{pa(i)}^{\mathcal{H}}
\end{equation*}
Note that the positional embeddings are initialized and then projected to hyperbolic space following~\tbref{tab:hyperbolic-operations}, while the concatenation is as described in~\secref{sec:hyperbolic-operations}. $\mathbf{x}_{i}^{0, \mathcal{H}}$ is the initial concept feature obtained following \secref{subsec:initial-concept-features}.

\vspace{.3em}

\mypar{Feature Transformation.}
At layer $l$, we transform the embedding vectors produced by the previous layer $\mathbf{x}_{i}^{\ell-1, H}$ to message $\mathbf{h}_{i}^{\ell, H}$ by applying a hyperbolic linear transformation:
\begin{equation*}
\mathbf{h}_{i}^{\ell, \mathcal{H}}=\left(W^{\ell} \otimes^{K} \mathbf{x}_{i}^{\ell-1, \mathcal{H}}\right) \oplus^{K} \mathbf{b}^{\ell}
\end{equation*}
where $\otimes^{K}$ and $\oplus^{K}$ denotes multiplication and addition in hyperbolic space $\mathcal{H}$ with curvature $K$.

\vspace{.3em}

\mypar{Neighborhood Aggregation.}
The neighborhood aggregation encapsulates neighboring features to update the center node.
To enable an importance-weighted aggregation and for the simplicity to reuse Euclidean operations to derive the attention scores, we firstly apply a logarithmic mapping to project the messages to a tangent space.
Let $i$ be the center node and $j$ be one of its neighbor nodes, we compute an attention weight $w_{ij}$ by applying an Euclidean MLP to the concatenated tangential representations of the two messages following:
\begin{equation*}
w_{i j} =\sigma_{j \in \mathcal{N}(i)}\left(\operatorname{MLP}\left(\log _{\mathbf{o}}^{K}\left(\mathbf{h}_{i}^{\mathcal{H}}\right) \| \log _{\mathbf{o}}^{K}\left(\mathbf{h}_{j}^{\mathcal{H}}\right)\right)\right).
\end{equation*}
where $\sigma$ is a softmax function over all neighbors $\mathcal{N}(i)$.
The center node embedding is thus obtained by a weighted sum of the neighboring tangential embeddings.
Finally, we apply an exponential mapping to project the aggregated tangential center node embedding to the hyperbolic model $\mathcal{H}$ as:
\begin{equation*}
\mathrm{AGG}^{K}\left(\mathbf{h}^{\mathcal{H}}_{i}\right) =\exp _{\mathbf{h}_{i}^{\mathcal{H}}}^{K}\left(\sum_{j \in \mathcal{N}(i)} w_{i j} \log _{\mathbf{h}_{i}^{\mathcal{H}}}^{K}\left(\mathbf{h}_{j}^{\mathcal{H}}\right)\right).
\end{equation*}
Note that for a better local hierarchical approximation, an \textit{independent} local tangent space is created for each center node $i$ during the neighborhood aggregation, instead of using the tangent space of the hyperbolic origin (\ie using $\exp _{\mathbf{h}_{i}^{\mathcal{H}}}^{K}$ and $\log _{\mathbf{h}_{i}^{\mathcal{H}}}^{K}$ instead of $\exp _{\mathbf{o}}^{K}$ and $\log _{\mathbf{o}}^{K}$).
The curvature $K$ of a hyperbolic model can either be a fixed number or a learnable parameter, where our experiments show that learned $K$ tends to yield better performance.
The update rule of the embedding of node $i$ can thus be defined as:
\begin{equation*}
\mathbf{x}_{i}^{\ell, \mathcal{H}}=\sigma(\mathrm{AGG}^{K_{\ell-1}}\left(\mathbf{h}^{\ell-1, \mathcal{H}}_{i}\right)),
\end{equation*}
and we concatenate the updated node embedding with taxonomy-specific features and use as input for next layer.
Finally we obtain the ego graph representation using the finalized node embeddings via a weighted readout function for the 1-hop neighbor nodes. Given $G$ as 1-hop ego graph, $pr_{a_i}(j)$ as node $j$'s relative positions (parent, child or self) related to center node $a_i$, $\alpha_{pr_{a_i}(j)}$ as the weight for node-type, then the concept representation for anchor node $a_i$ is:
\begin{equation*}
\mathbf{o}_{a_i}=\sum_{j \in G} \frac{\log \left(1+\exp \left(\alpha_{pr_{a_i}(j)}\right)\right)}{\sum_{j^{\prime} \in G} \log \left(1+\exp \left(\alpha_{pr_{a_i}(j^{\prime})}\right)\right)} \mathbf{x}_{i}^{\ell, \mathcal{H}}.
\end{equation*}

\subsection{Matching Module}
\label{subsec:match-function}

Given the initial concept features $x_{q_{i}}$ of a query concept $q_i$, we obtain the query concept representation $\mathbf{o}_{q_i}$ by projecting $x_{q_{i}}$ to the hyperbolic space $\mathcal{H}$ using the exponential mapping function (if $x_{q_{i}} \in \mathcal{E}$) or hyperbolic model transformation (if $x_{q_{i}}$ is in other hyperbolic models other than $\mathcal{H}$) defined in \secref{sec:hyperbolic-operations}. Note that the hyperbolic spaces used to obtain the anchor and query concept representations need to be consistent.

After obtaining the hyperbolic embedding representation for each query concept 
$
\mathbf{o}_{q_i} \in \mathcal{H}
$
and each anchor concept
$
\mathbf{o}_{a_i} \in \mathcal{H}
$,
$\mathbf{o}_{q_i}$ and $\mathbf{o}_{a_i}$ are concatenated with hyperbolic operations, and then we feed the concatenated vector to an HNN.
We construct hyperbolic multi-layer perceptron (MLP) based on the operations defined in~\cite{Ganea2018HyperbolicNN}, and a one-layer HNN is defined as:
\begin{equation*}
f^{\mathrm{HNN}}\left(x\right)=\varphi^{\otimes^{K}}(M \otimes^{K} x \oplus^{K} b)
\end{equation*}
where $M \in \mathcal{R} ^{m \times n}$ and $b \in \mathcal{H} ^{m}$ are learnable parameters.
Since $b$ lies in a hyperbolic space, its update during training needs to be calibrated to remain in the proper manifold. 
$\varphi^{\otimes}$ is the element-wise non-linearity, where $\otimes^{K}$ and $\oplus^{K}$ denotes multiplication and addition in hyperbolic space, respectively, under the curvature $K$.
Note that HNN is equivalent to a Euclidean MLP if $K$ is set to 0, \ie the embedding space is not \textit{curved}.



\subsection{Learning and Inference}
\label{subsec:training-learning}

We train the~\modelname framework with a metric learning paradigm by utilizing the existing taxonomies as the training resources.

\vspace{.3em}

\mypar{Training Data Construction.} 
The data pairs that are used to train the matching module is generated in a self-supervised manner following \citet{Shen2020TaxoExpanST}.
We only consider \textit{exact} parent-child node pairs on the seed taxonomy $\mathcal{T}^0$ as the positive samples, \ie there exists a \textit{direct} edge $\langle a_i, q_i \rangle$.
For each query node $q_i$, we randomly sample $N$ other nodes (without its immediate children) on the seed taxonomy to form negative training instances $\langle n_i^1, q_i \rangle, \langle n_i^2, q_i \rangle, ..., \langle n_i^N, q_i \rangle$.
Anchoring at node $q_i$, the positive and negative samples form a single group of training instances $\mathbf{X}_{i}=\left\{\left\langle a_i, q_i\right\rangle,\left\langle n_{i}^{1}, q_i\right\rangle, \ldots,\left\langle n_{i}^{N}, q_i\right\rangle\right\}$.
We repeatedly apply this operation on each edge of the seed taxonomy to construct our training data $\mathbb{X}=\left\{\mathbf{X}_{1}, \ldots, \mathbf{X}_{\left|\mathcal{E}^{0}\right|}\right\}$.

\vspace{.3em}



\vspace{.3em}

\mypar{Learning Objective.}
We adopt InfoNCE loss \cite{oord2018representation} as the main training objective:
\begin{equation*}
\mathcal{L}(\Theta)=-\frac{1}{|\mathbb{X}|} \sum_{\mathbf{X}_{i} \in \mathbb{X}}\left[\log \frac{f\left(a_{i}, q_{i}\right)}{\sum_{\left\langle n_{i}^{j}, q_{i}\right\rangle \in \mathbf{X}_{i}} f\left(n_{i}^{j}, q_{i}\right)}\right]
\end{equation*}
where $j \in [0,1,2,...,N]$ and $n_{i}^{0}$ is the positive sample $a_i$.
The InfoNCE loss is essentially a cross entropy loss which identifies the positive pairs (items in the numerator) among all the possible candidates (items in the denominator).

\vspace{.3em}

\mypar{Inference.}
During the inference time, for each new query concept (unseen from the seed taxonomy) $q_i$, we compute the matching scores between the query concept $q_i$ and every candidate anchor nodes $a_{candidate} \in \mathcal{N}^0$ in the existing taxonomy $\mathcal{T}^0$. We then rank these anchor nodes by the matching score to create a ranked list of length $\lvert \mathcal{N}^0 \rvert$ for deciding where to attach such new concept. 


\section{Experiments}
\label{sec:eval}

We evaluate the~\modelname and its variants on four large-scale real-world taxonomies utilized by~\citet{Shen2020TaxoExpanST} and~\citet{zhang2021tmn}.

\subsection{Experimental Setup}

\mypar{Datasets.}
\label{subsec:datasets}
Following \citet{Shen2020TaxoExpanST,zhang2021tmn},
we take WordNet 3.0 and 1000 domain-specific concepts defined in SemEval-2016 Task 14 Benchmark dataset \cite{jurgens2016semeval}, where only hypernym-hyponym relations are considered.
WordNet thereof is separated into the verb version \textit{WordNet-Verb} and the noun version \textit{WordNet-Noun}.
We also use subgraphs of the Field-of-Study Taxonomy in Microsoft Academic Graph \cite{sinha2015overview} containing descendants of ``psychology'' and ``computer science'' node and refer as \textit{MAG-PSY} and \textit{MAG-CS}.

\begin{table}[h]
\begin{center}
\resizebox{\linewidth}{!}{
\begin{tabular}{l@{\hskip9pt} | 
c@{\hskip9pt}| c@{\hskip9pt}|c@{\hskip9pt}c}
\toprule

Dataset & \# Nodes & \# Edges & Depth  \\
\midrule

WordNet-Verb & 13,936 & 13,408 & 13 \\
WordNet-Noun & 83,073 & 76,812 & 20 \\
MAG-PSY & 23,187 & 30,041 & 6 \\
MAG-CS       & 24,754 & 42,329 & 6 \\
\bottomrule
\end{tabular}
}
\caption{Dataset statistics.}
\label{table:dataset}
\end{center}
\end{table}

More detailed statistics of each dataset are in~\tbref{table:dataset}.
For each dataset, 1000 leaf nodes are randomly sampled as query nodes as the validation set, and another 1000 leaf nodes form the test set.
Since these validation and testing nodes are all leaf nodes, only minimum changes are required to make the remaining taxonomy still a valid one without introducing non-existed edges.
For WordNet-Verb and WordNet-Noun, we generate the initial concept features following \secref{subsec:initial-concept-features}. We assume each concept has only one name and we induce the concept name from the WordNet synset name.
For MAG-PSY/CS, we use 250-d in-domain concept name word embeddings provided by \citet{Shen2020TaxoExpanST} trained using skipgram model on paper abstract corpus. Since we do not have access to the source profile information, we cannot obtain initial concept features as designed in~\secref{subsec:initial-concept-features}. As a result, we cannot run two RoBERTa-base baseline models introduced in the following section on the MAG-PSY/CS dataset.


\mypar{Evaluation Metrics.}
We follow prior studies \cite{zhang2021tmn,Shen2020TaxoExpanST,manzoor2020expanding} to report 
several widely-used ranking metrics: \emph{MeanRank (MR)}, \emph{Mean Reciprocal Rank (MRR)},\footnote{We report the MRR numbers scaled by 10 following previous works to amplify the performance difference.} \emph{Recall @ K} and \emph{Precision @ K}.

\paragraph{Baseline Models.}
We compare~\sysname with the following strong baseline models:
\begin{itemize}[leftmargin=*]
\setlength\itemsep{0em}
    \item \textbf{RoBERTa-base Zero-shot}: we use RoBERTa-base as feature extractor to obtain initial embeddings as described in \secref{subsec:initial-concept-features} without fine-tuning
    \item \textbf{RoBERTa-base FT}: the above design but update the LM's parameters
    \item \textbf{Hyperbolic MLP}: we concatenate initial features of query and anchor concepts and feed into a two-layer hyperbolic MLP
    \item \textbf{GCN} \cite{kipf2016semi}: \modelname's design but use Euclidean GCN to update node embeddings in ego graph of the anchor concept, use fastText to obtain initial concept features, and use Euclidean MLP as the matching module
    \item \textbf{GAT} \cite{velivckovic2017graph}: the above method but use GAT to update node embeddings.
\end{itemize}

\noindent We compare~\modelname with the following state-of-the-art taxonomy expansion frameworks:
\begin{itemize}[leftmargin=*]
\setlength\itemsep{0em}
    \item \textbf{TaxoExpan} \cite{Shen2020TaxoExpanST} uses GCN and GAT to get node embeddings of ego networks of anchor nodes and average all node embeddings to get anchor concept representation. But the ego network only includes direct children and parent of the anchor concept. They also inject relative positional embeddings to GNN.
    \item \textbf{ARBORIST} \cite{manzoor2020expanding} combines global and local taxonomic information to explicitly model heterogeneous and unobserved edge semantics. 
    \item \textbf{TMN} \cite{zhang2021tmn} uses auxiliary scorers to capture various fine-grained signals including query to hypernym and query to hyponym semantics and introduces a channel-wise gating mechanism to retain task-specific information.
\end{itemize}

\subsection{Experimental Results}

\begin{table*}[ht]
\begin{center}
{
\small
\setlength\tabcolsep{2pt}
\begin{tabular}{l | 
r c|r
rr|r
rr|r c|
rrr|rrr}
\toprule

\multirow{3}{*}{Model} & \multirow{2}{*}{MR $\downarrow$}  & \multirow{2}{*}{MRR $\uparrow$} & \multicolumn{3}{c|}{Recall \% $\uparrow$} &  \multicolumn{3}{c|}{Precision \% $\uparrow$} & \multirow{2}{*}{MR $\downarrow$}  & \multirow{2}{*}{MRR $\uparrow$} & \multicolumn{3}{c|}{Recall \% $\uparrow$} &  \multicolumn{3}{c}{Precision \% $\uparrow$} \\
& & & @1 & @5 & @10 & @1 & @5 & @10 & & & @1 & @5 & @10 & @1 & @5 & @10   \\ \cmidrule{2-17}
& \multicolumn{8}{c|}{WordNet-Verb \footnotesize{(Candidates \#: 11,936)}} & \multicolumn{8}{c}{WordNet-Noun \footnotesize{(Candidates \#: 81,073)}} \\
\midrule
ARBORIST 
    \wordnetverbARBORIST \wordnetnounARBORIST \\
TaxoExpan 
    \wordnetverbTaxoExpanGCN \wordnetnounTaxoExpanGCN \\
TMN 
    \wordnetverbTMN \wordnetnounTMN \\
\midrule
RoBERTa-base 0-shot
    \wordnetverbRoberta \wordnetnounRoberta \\
RoBERTa-base FT 
    \wordnetverbRobertaFT \wordnetnounRobertaFT \\
Hyperbolic MLP 
    \wordnetverbHyperbolicMLP \wordnetnounHyperbolicMLP \\
GCN 
    \wordnetverbGCN \wordnetnounGCN \\
GAT 
    \wordnetverbGAT \wordnetnounGAT \\ \midrule
\sysname 
    \wordnetverbOursBold \wordnetnounOurs \\

\toprule

& \multicolumn{8}{c|}{MAG-PSY \footnotesize{(Candidates \#: 21,187)}} & \multicolumn{8}{c}{MAG-CS \footnotesize{(Candidates \#: 22,754)}} \\ 
\midrule
ARBORIST 
    \magpsyARBORIST \magcsARBORIST \\ 
TaxoExpan 
    \magpsyTaxoExpan \magcsTaxoExpan \\
TMN 
    \magpsyTMN \magcsTMN \\
\midrule
Hyperbolic MLP  
    \magpsyHyperbolicMLP \magcsHyperbolicMLP \\
GCN 
    \magpsyGCN \magcsGCN \\
GAT 
    \magpsyGAT \magcsGAT \\ \midrule
\sysname 
    \magpsyOurs \magcsOurs \\

\bottomrule
\end{tabular}
}

\caption{
Overall experimental results. Directions (pointing up or down) of arrows indicate better performance of the metrics. MRR metrics are scaled by 10 to amplify the performance difference.
}
\label{table:overall}
\end{center}
\end{table*}

The overall experimental results are shown in \tbref{table:overall}. We introduce our implementation details and hyperparameter settings in \appref{appendix:impdetails}. 

Among ARBORIST, TaxoExpan and TMN, TMN achieves the strongest result consistently. Note that TMN is trained on taxonomy completion task and only perform inference on taxonomy expansion task. Anchor node representations are learned coupled with different potential children of the query concept which provides fine-grained signals.
TaxoExpan performs better than ARBORIST showing the importance of neighborhood information.
Experiments using RoBERTa-base on two WordNet datasets indicate that RoBERTa language model falls drastically behind in this context-independent task. Since the profile sentences are very short and the task is more rely on commonsense rather than context understanding, language models cannot benefit from contextualized representation, which consolidates the observation by \citet{liu2020towards}. 
We can observe Hyperbolic MLP is worse than GNN models since it does not use neighborhood profile information. Hyperbolic MLP outperforms ARBORIST with a large margin on all datasets.
The comparison between GCN and GAT indicates that attentive aggregation is more helpful with the sparse neighborhood representation. If we compare \modelname with GCN and GAT, we can observe that the expressiveness of the hyperbolic space leads to a large performance increase (9.5\% and 6.9\% recall@10 increase on MAG-PSY and WordNet-Verb and MRR scaled by 10 increase ranging from 0.013 to 0.076). Overall, \modelname consistently outperforms all models across four datasets except MR for WordNet-Noun. 

\begin{table}[t]
\begin{center}
{\small
\setlength\tabcolsep{4pt}
\begin{tabular}{c|l|
Hc|H
Hc|c
HHr r
cccccc}
\toprule
\multirow{2}{*}{\#} &\multirow{2}{*}{Model} & \multirow{2}{*}{MeanRank $\downarrow$} & \multirow{2}{*}{MRR $\uparrow$} & \multicolumn{3}{c|}{Rec $\uparrow$} &  \multicolumn{3}{c}{Prec $\uparrow$} \\

 & & & & @1 & @5 & @10 & @1 & @5 & @10   \\
\midrule
1 & w/o trainable curvature \wordnetverbCurvefixBest \\
2 & \poincare i/o Lorentz model 
    & 473.5 & 0.494 & 13.0 & 30.6 & 39.8 & 13.0 & 6.1 & 4.0 \\ 
3 & fastText i/o \poincare GloVe \wordnetverbfasttext \\
\midrule
4 & anchor + parent + children \wordnetverbOursOneHop \\ 
5 & \#4 + anchor's ancestors \wordnetverbOursOneHopAncestors \\ 
6 & \#5 + anchor's descendants \wordnetverbOurs\\ 
7 & \#6 + anchor's siblings \wordnetverbOursTwoHops \\ 
\midrule
8 & w/o Relative Pos Emb \wordnetverbOursAbsPos \\
9 & w/o Absolute Pos Emb \wordnetverbOursRelPos \\
10 & w/o both Positional Emb \wordnetverbOursNoPos \\
\midrule
 & \sysname \wordnetverbOurs \\
\bottomrule
\end{tabular}
}
\caption{Experimental results for ablation studies on WordNet-Verb. By default, we use trainable curvature, Lorentz hyperbolic model, \poincare GloVe as initial word embedding, 2-hop computational graph without anchor's sibilings, with both relative and absolute position embedding. ``i/o'' means ``instead of'', ``w/o'' means ``without''.
}
\label{table:ablation}
\end{center}
\end{table}


To further help understand the contribution of different incorporated techniques,
we present a series of ablation study results in \tbref{table:ablation}. Specifically, we have the following observations:

According to lines 1-3, the trainable curvature learns fine-grained suitable manifold setting and lead to almost 2\% recall@10 improvement (lines 1-3). 
Replacing the default Lorentz model with \poincare model notably hinders the performance which can be explained by Lorentz model's numerical stability since the distance function of the \poincare model contains fraction \cite{Chami2019HyperbolicGC,Peng2021HyperbolicDN}.
We replace \poincare GloVe initial word embedding with fastText in line 3 and the result shows that \poincare GloVe contains better feature for our task.

We explore different choices of neighborhood aggregation in lines 4-7. 
We observe that 2-hop neighborhood aggregation leads to improvement over 1-hop in terms of recall@10 and precision@1 (line 5). Adding descendant of the anchor node supports with better characterization of nodes (line 6). However, we observe a noticeable drop when we further add sibling nodes into the aggregation neighborhood (line 7). The potential reason is that the sibling nodes are very diverse, and thus are not closely related to the anchor node. 

In lines 8 to 10, we investigate the effect of positional embeddings. A larger performance drop is caused if we remove relative position embeddings (line 8), in comparison to a lesser drop when removing the absolute position embedding (line 9). We hypothesize that the absolute position embedding (depth information) is provided implicitly in the ego graph by edges among events. Line 10 shows that both embeddings are essential to boost the performance by almost 4\% gain in recall@10.






\section{Related Works}
\label{sec:relatedworks}

Our work is connected to two lines of research. 


\paragraph{Taxonomy Expansion}

Taxonomy expansion task fits in real-world application scenario that 
automatically attach new concepts or terms into a human curated seed taxonomy \cite{vedula2018enriching}. 
Traditional methods leverage pre-defined patterns to extract hypernym-hyponym pairs for taxonomy expansion \cite{nakashole2012patty, jiang2017metapad, agichtein2000snowball}.
Some works use external data and expand taxonomy in a specific domain. For example, \citet{toral2008named} use Wikipedia named entities to expand WordNet, \citet{wang2014hierarchical} use query logs to expand search engine category taxonomy. 
Some works expand a generic taxonomy without using external resources. 
For example, \citet{shwartz2016improving} encode taxonomy traversal paths to seize on the dependency between concepts, \citet{Shen2020TaxoExpanST} use a GNN model that handles this task, ARBORIST \cite{manzoor2020expanding} produces concept representations using signals from both edge semantics and surface forms of concepts. STEAM \cite{yu2020steam} formulates the taxonomy expansion task as a mini-path-based prediction task and introduces a co-training process for semi-supervised learning. Recently, \citet{zhang2021tmn} propose the taxonomy completion task in which the new concept can be inserted between existing concepts on taxonomy.
\citeauthor{zhang2021tmn} also introduce the TMN model whose auxiliary scorers capture different fine-grained signals. 
Comparing with these methods using Euclidean space, \modelname uses hyperbolic representation learning to provide feature space with low distortion especially for lower-level concepts on taxonomies.


\paragraph{Hyperbolic Representation Learning} 

\citet{nickel2017poincare} present an efficient algorithm to learn embeddings in a supervised manner based on Riemannian optimization and shows it performs well on link prediction task even with a smaller dimension.
\citet{Ganea2018HyperbolicNN} presents common neural network operations in hyperbolic space and \citet{Liu2019HyperbolicGN} extends GNN operations to Riemannian manifolds with differentiable exponential and logarithmic maps. Most related to our work, \citet{Chami2019HyperbolicGC} derives Graph Convolutional Neural Network (GCN)'s operations in the Lorentz model of hyperbolic space. 
Hyperbolic representation learning is broadly applied to lexical representations \cite{dhingra2018embedding, tifrea2018poincare, zhu-etal-2020-hypertext}, organizational chart induction \cite{chen2019embedding}, hierarchical classification \cite{lopez-strube-2020-fully,chen2020hyperbolic}, knowledge association \cite{sun2020knowledge}, knowledge graph completion \cite{wang2021mixed,balazevic2019multi} and event prediction \cite{suris2021hyperfuture}.
A more comprehensive summarization is given in a recent survey by \citet{Peng2021HyperbolicDN}.

There are studies that leverage hyperbolic representation learning to perform taxonomy extraction from text, which are connected to this work.
Such studies use \poincare embeddings trained by hypernymy pairs extracted by lexical-syntactic patterns \cite{hearst1992automatic} to infer missing nodes \cite{le2019inferring} and refine preexisting taxonomies \cite{alyetal2019every}. The patterns suffer from missing and incorrect extractions, and are dedicated to capturing hypernymy relations between nouns. 
Hence, only terms that are recognizable by the designed patterns are able to be attached to the taxonomy.
These works solely rely on graph structures of the taxonomy to obtain hyperbolic embeddings of known concepts, and 
cannot handle emerging, unseen concepts using their profile information.
This is one of the problems that are addressed in this work.



\section{Conclusion and Future Work}
\label{sec:conclusion}
We present \modelname, a taxonomy expansion model which better preserves the taxonomical structure in an expressive hyperbolic space. We use an HGNN to incorporate neighborhood information and positional features of concepts, as well as profile features that are essential to jump-start zero-shot concept representations. Experimental results on WordNet and Microsoft Academic Graph taxonomies show that \modelname performs better than its Euclidean counterparts and consistently outperforms state-of-the-art taxonomy expansion models. In the future, we plan to 
extend \modelname for inducing dynamic taxonomies \cite{zhu2021learning} and taxonomy alignment \cite{sun2020knowledge}.

\section*{Acknowledgments}

Many thanks to Liunian Harold Li, Fan Yin, I-Hung Hsu, Rujun Han and Shuowei Jin for contribution, discussion and internal reviews, to lab members at the PLUS lab and UCLA-NLP for suggestions, and to the anonymous reviewers for their feedback.
This material is based on research supported by DARPA under agreement number FA8750-19-2-0500. The U.S. Government is authorized to reproduce and distribute reprints for Governmental purposes notwithstanding any copyright notation thereon. The views and conclusions contained herein are those of the authors and should not be interpreted as necessarily representing the official policies or endorsements, either expressed or implied, of DARPA or the U.S. Government. 

\section*{Ethical Considerations}
This work does not present any direct societal consequence. The proposed method aims at improving representation learning to support automated expansion of taxonomies.
We believe this study leads to intellectual merits that benefit from automated knowledge acquisition for constructing knowledge representations with complex or sparse structures.
It could also potentially lead to broad impacts, as the obtained taxonomical knowledge representations can support various knowledge-driven tasks.
It is important to note that the precision of top taxonomy expansion predictions is still not high even with the state-of-the-art method, so human validation is needed before the taxonomy generated by the automated method is used in real-world applications.
All experiments are conducted on open datasets.


\bibliography{ma}
\bibliographystyle{acl_natbib}

\clearpage

\appendix


\section{Graph Convolutional Networks}
\label{appendix:gcn}
Graph convolutional network (GCN)~\cite{kipf2016semi} is a widely-used variant of graph neural network.
GCN defines one hop of graph message passing as a combination of the feature transformation and the neighborhood aggregation at a single layer $l$.
The input feature transformation is defined as:
\begin{equation*}
\mathbf{h}_{i}^{\ell, \mathcal{E}} =W^{\ell} \mathbf{x}_{i}^{\ell-1, \mathcal{E}}+\mathbf{b}^{\ell}
\end{equation*}
where $\mathcal{N}(i)=\{j:(i,j)\in \mathcal{E}\}$ is a set of neighboring nodes of node $i$, $W^{\ell}$ and $b^{\ell}$ are learnable weight and bias parameters for layer $l$.
The neighborhood aggregation is then defined as:
\begin{equation*}
\mathrm{AGG}^{0}\left(\mathbf{x}^{\ell, \mathcal{E}}\right)_{i}
=\sigma\left(\mathbf{h}_{i}^{\ell, \mathcal{E}}+\sum_{j \in \mathcal{N}(i)} w_{i j} \mathbf{h}_{j}^{\ell, \mathcal{E}}\right)
\end{equation*}
where $w_{ij}$ denotes the scores for a weighted aggregation, \ie how important node $j$ is for node $i$, and $\sigma$ is a non-linear activation function.
By cascading multiple layers of GCN, the message can be propagated over several hops of neighborhoods.
The node embeddings in the graph are being updated during the training process.
Notice that the superscript $0$ in the above equation denotes the $0$-curved space, \ie, the aggregation is performed in a Euclidean space. 

\section{Implementation Details}
\label{appendix:impdetails}

All the models in this work are trained on a single Nvidia A100 GPU\footnote{https://www.nvidia.com/en-us/data-center/a100/} on a Ubuntu 20.04.2 operating system.
The hyperparameters for each model are manually tuned against different datasets, and the checkpoints used to evaluate are selected by the best performing ones on the development set.

Our entire code-base is implemented in PyTorch.\footnote{https://pytorch.org/}
The implementations of the transformer-based models are extended from the huggingface\footnote{https://github.com/huggingface/transformers}~code base~\cite{wolf-etal-2020-transformers}.
The implementations of the models compared with, \ie TMN, TaxoExpan and ARBORIST, are obtained and adapted from the original author released code repositories.

\subsection{Hyper-parameters}

We introduce the hyper-parameters used throughout this work and the searching bounds for the manual hyper-parameter tuning in~\tbref{table:search}.

We set burnin epoch number to 20 during which we use 1e-5 learning rate, after the burnin epochs, the learning rate is 1e-3 with ReduceLROnPlateau scheduler with 10 patience epochs. For each positive sample, we generate 31 negative samples. Dimension for anchor concept representation (output dimension of HGNN) is set to 100. We use two GNN layers by default. We use stochastic Riemannian Adam optimizer \cite{geoopt2020kochurov,nickel2017poincare}. For absolute and relative positional embedding, we use 50 dimensions by default. We use MRR of the validation set as the metric to monitor for an early stop.


\begin{table}[t]
\centering
\footnotesize
\begin{tabular}{ccccc}
    \toprule
    \textbf{Type} & \textbf{Batch Size} & \textbf{Initial LR}  \\
    \midrule
    \textbf{Bound (lower--upper)} & 8-128 & $1 \times 10^{-2}$--$1 \times 10^{-6}$ \\
    \midrule
    \textbf{Number of Trials} & 2--4 & 2--3 \\
    \bottomrule
\end{tabular}
\caption{
\textbf{Search bounds:} for the hyperparameters of all the models.
}
\label{table:search}
\end{table}

\end{document}